\documentclass[conference]{IEEEtran}
\IEEEoverridecommandlockouts
\usepackage{cite}
\usepackage{amsmath,amssymb,amsfonts}
\usepackage{algorithmic}
\usepackage{graphicx}
\usepackage{textcomp}
\usepackage{xcolor}
\usepackage{booktabs}
\usepackage{multirow}
\usepackage{multicol}

\def\BibTeX{{\rm B\kern-.05em{\sc i\kern-.025em b}\kern-.08em
    T\kern-.1667em\lower.7ex\hbox{E}\kern-.125emX}}
\begin{document}

\title{Establishing a Stronger Baseline \\for Lightweight Contrastive Models
\thanks{This research is supported by National Natural Science Foundation of China (Grant No.62276154), Research  Center for Computer Network (Shenzhen) Ministry of Education, Beijing Academy of Artificial Intelligence (BAAI), the Natural Science Foundation of Guangdong Province (Grant No.  2023A1515012914), Basic Research Fund of Shenzhen City (Grant No. JCYJ20210324120012033 and JSGG20210802154402007), and the Major Key Project of PCL for Experiments and Applications (PCL2021A06).}
}

% \author{Wenye Lin$^{1,2*}$\thanks{$^*$ Equal contributions.}, Yangming Li$^{1*}$\samethanks, Lemao Liu$^1$, Shuming Shi$^1$, Hai-Tao Zheng$^{2,3\dagger}$\thanks{$^{\dagger}$ Corresponding author.}}
% {$^1$Tencent AI Lab \\
%     $^2$Shenzhen International Graduate School, Tsinghua University \\
%     $^3$Pengcheng Laboratory}

\author{\IEEEauthorblockN{Wenye Lin$^{1}$, Yifeng Ding$^{1}$, Zhixiong Cao$^{1}$, Hai-Tao Zheng$^{1,2\dagger}$\thanks{$^{\dagger}$ Corresponding author.}}

\IEEEauthorblockA{$^1$Shenzhen International Graduate School, Tsinghua University, China}
\IEEEauthorblockA{$^2$Peng Cheng Laboratory}
\IEEEauthorblockA{\texttt{\{lwy20,dingyf20,caozx21\}@mails.tsinghua.edu.cn}}
\texttt{zheng.haitao@sz.tsinghua.edu.cn}
}

% \author{\IEEEauthorblockN{1\textsuperscript{st} Wenye Lin}
% \IEEEauthorblockA{\textit{Shenzhen International Graduate School} \\
% \and
% \and
% \and

% }

\maketitle

\begin{abstract}
Recent research has reported a performance degradation in self-supervised contrastive learning for specially designed efficient networks, such as MobileNet and EfficientNet. A common practice to address this problem is to introduce a pretrained contrastive teacher model and train the lightweight networks with distillation signals generated by the teacher. However, it is time and resource consuming to pretrain a teacher model when it is not available. In this work, we aim to establish a stronger baseline for lightweight contrastive models without using a pretrained teacher model.
Specifically, we show that the optimal recipe for efficient models is different from that of larger models, and using the same training settings as ResNet50, as previous research does, is inappropriate. Additionally, we observe a common issue in contrastive learning where either the positive or negative views can be noisy, and propose a smoothed version of InfoNCE loss to alleviate this problem. As a result, we successfully improve the linear evaluation results from 36.3\% to 62.3\% for MobileNet-V3-Large and from 42.2\% to 65.8\% for EfficientNet-B0 on ImageNet, closing the accuracy gap to ResNet50 with $5\times$ fewer parameters. We hope our research will facilitate the usage of lightweight contrastive models. \footnote{Our code is available at https://github.com/Linwenye/light-moco.}

\end{abstract}

\section{Introduction}
\label{sec:intro}

Self-supervised contrastive learning targets to learn the semantic representation of instances through pretraining on the instance discrimination pretext task. In this task, a model is trained to align augmented views from the same instance and push away from views of other instances in the representation space. Leading contrastive learning methods using large models have already shown superiority to supervised counterpart when fine-tuning the pretrained models on downstream tasks \cite{he2020momentum,grill2020bootstrap,chen2021exploring}. 
% Self-supervised contrastive learning targets at learning semantic representation of instances through pretraining on the instance discrimination pretext task, where a model is trained to align augmented views from the same instance, while pushing away from those of others in the representaion space. Notably, the leading contrastive methods utilizing large models are shown to surpass the supervised counterpart after finetuning on downstream tasks \cite{chen2020simple,he2020momentum,grill2020bootstrap,chen2021exploring}.

However, previous research reports that the performance of these methods suffers a cliff fall with the decrease of model size. For example, the supervised top-1 accuracy on ImageNet for ResNet-50/EfficientNet-B0 is 76.1\%/77.1\%, but the corresponding linear evaluation results applying MoCo-V2~\cite{chen2020improved} is 67.5\%/42.2\%, indicating a large gap. To rescue the performance of small models, a widely adopted practice is to leverage a large pretrained model to transfer the representational knowledge through knowledge distillation~\cite{fang2020seed,xu2021bag,gao2021disco}. In spite of the effectiveness, such a pretrained teacher model is not always available, especially in some domain-specific scenarios, and it is time-consuming and resource-intensive to pay extra cost to train one. Additionally, it may not be possible to simultaneously load the student and teacher models when memory resources are limited. Recently, Shi et al.~\cite{shi2022efficacy} attempt to train small self-supervised models without distillation signals, but the performance gap still exists. Furthermore, a pretrained large model is still required to guild the lightweight model to a better initialization. 
% Recently, \cite{shi2022efficacy} observe that small models are inclined towards over-clustering solutions, where augmented views from the same instance are tightly clustered, whilst other instances with similar semantics are not well clustered in the representation space. They then search various hyperparameter settings and improve the baseline empirically but the rationale is unclear according to the authors \cite{shi2022efficacy}. Besides, one of the most significant improvements still comes from the assistance from the teacher model at the early stage.
\begin{figure}
         \centering
         \includegraphics[width=0.45\textwidth]{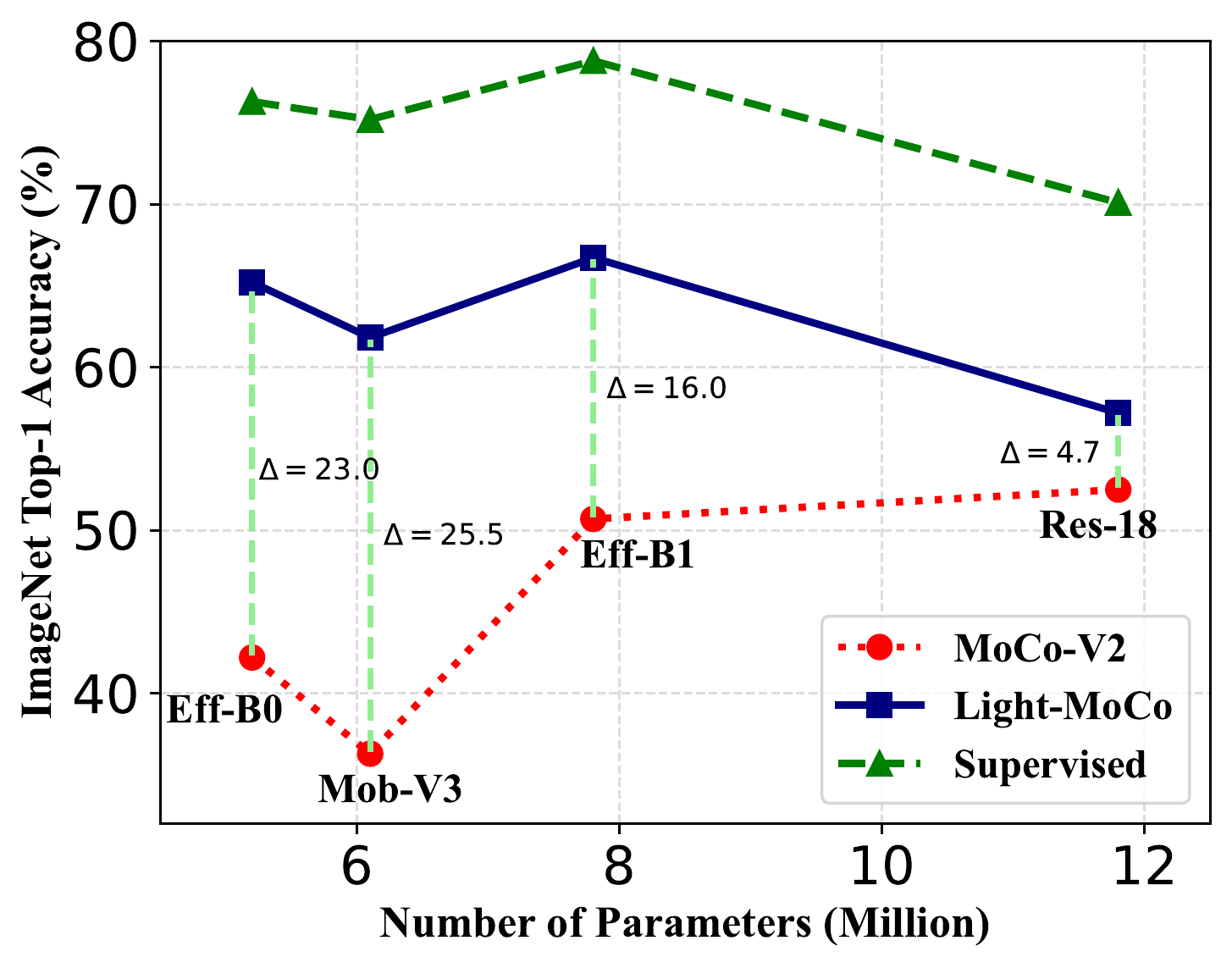}
         \caption{Linear probe top-1 accuracy of Light-MoCo (ours) and MoCo-v2 on ImageNet. The top line is the supervised counterpart. Our method dramatically boosts the performance based on MoCo-v2 without a teacher model. }
        \label{fig:comparison}
\end{figure}

In this work, we endeavor to establish a stronger baseline for lightweight contrastive models based on MoCo-V2 without resorting to a powerful teacher model. We show that while recent research directly adopts the default setting as the baseline for lightweight models, it is far from optimal in terms of performance~\cite{fang2020seed,xu2021bag,gao2021disco}. We benchmark a few representative self-supervised methods on ImageNet using the same training settings as ResNet50, and observe a wide existence of performance degradation for ResNet18 with these methods (Results are shown in Table~\ref{tab:imagenet}). After extensively exploring the best recipe for lightweight contrastive models, we largely boost their evaluation accuracy in downstream tasks. Among the explored training techniques, a larger learning rate and a wider MLP head contribute most to reduce the performance gap to large self-supervised models. To further boost its performance, We also alleviate the problem of noisy views (i.e., false positive views and false negative views) during self-supervised training\cite{kalantidis2020hard,huynh2022boosting} with SoftNCE, a simple loss generalizing the widely used InfoNCE loss to a smoothed one. Instead of aligning with a one-hot distribution in InfoNCE, our loss enhances the instance discrimination objective by matching a softened probability distribution. The proposed smoothing technique prevents models from over-confidence, thus is more robust to noisy labels and generalize better.

Our main contributions are summarized as below:
% Mathematically, the training objective can be formulated as a cross entropy loss, by assigning probability 1 to align two augmented views for an instance, and probability 0 to all the other instances. We call the traditional InfoNCE loss as a ``hard" one.

% Which implies that the model simply learns a pattern to distinguish whether the two given views are generated from the same sample. There are two directions to deal with this problem. One is to apply heavier data augmentation, to make it less differentiable. less strict, strictly matching 

%  We investigate various weight assigning strategies and find it works best to distribute weight according to embedding distance of instances. 
% This goes against our anticipation that intra-class instances should be clustered together.
\begin{itemize}
    % \item We fill the gap to benchmark representative state-of-the-art contrastive learning methods of ResNet-18 on ImageNet. We hope this will accelerate related researches on light-weight contrastive learning.
    % \item Representative state-of-the-art contrastive learning methods typically base on large models, e.g., ResNet-50. In this work, we fill the gap to benchmark these methods with the lightweight model ResNet-18 on ImageNet.

    \item We empirically show that the default training setting for ResNet50 is suboptimal for lightweight models in contrastive learning, which is inappropriately adopted as a weak baseline in previous research \cite{fang2020seed,gao2021disco,xu2021bag}. We thus provide a better recipe for training lightweight contrastive models.
    \item We propose a novel loss, SoftNCE to mitigate the side effect from noisy views, improving the robustness and generalization of the models.
    \item We eventually show the feasibility to train lightweight contrastive models without relying on a powerful teacher model. For example, the linear evaluation accuracy of EfficientNet-B0 is improved from 42.2\% to 65.8\%, which is pretty close to ResNet-50 (67.4\%), with only 16.3\% of the number of parameters.

\end{itemize}

\begin{table}[]
    \centering
        \caption{20-NN and linear probe benchmarking results of ResNet-18 on ImageNet-1K with 512-batch self-supervised training, compared to ResNet-50 counterpart reported in their original paper. All the self-supervised learning methods are trained for 200 epochs. }
    \label{tab:imagenet}
    \begin{tabular}{lcccc}
    \toprule
    \multirow{2}{*}[-2pt]{Method}& \multicolumn{2}{c}{ResNet-50}& \multicolumn{2}{c}{ResNet-18}
     \\\cmidrule(lr){2-3}\cmidrule(lr){4-5}
     &  20-NN &Top-1 &  20-NN &Top-1\\
     \midrule
      MoCo-V2~\cite{chen2020improved} &-& 67.5& 44.6 & 52.5\\
      SimCLR-V2~\cite{chen2020big} &-&71.7& 43.8 & 51.9 \\
      SimSiam~\cite{chen2020simple}  &-&70.0& 23.0 & 32.7\\
      BYOL~\cite{grill2020bootstrap}  & 59.2&69.3&44.8 & 52.6\\ 
      PCL-V2~\cite{li2021prototypical} &- &67.6&44.2 & 52.1 \\
      MSF~\cite{koohpayegani2021mean} &64.9 &72.4&22.1 & 29.7\\
    %   ISD~\cite{}  &62.0 & 69.8&45.8 & 52.8 \\
      \midrule
      \textit{Supervised} & -& \textit{76.1}&-&\textit{70.2} \\
    %   \textit{ResNet-50} & \textit{57.3} & \textit{67.5} \\
    %   MoCo v2 (repr.) & 46.0  & 53.6\\
    %   Light-MoCo  & \textbf{48.7} & \textbf{57.2} \\
      \bottomrule
    \end{tabular}
\end{table}

\section{Related Work}
\subsection{Self-supervised Representation Learning}
There is a surge in research on self-supervised contrastive learning in recent years. Dosovitskiy et al.~\cite{dosovitskiy2014discriminative} first introduce the instance-level discrimination pretext task into self-supervised learning, which encourages a model to learn invariance of augmented views from an instance. Following research continuously pushes the limits and even surpasses the supervised counterpart in downstream tasks~\cite{he2020momentum,chen2020improved,chen2020big}. The commonly adopted InfoNCE loss function is as follows: 
\begin{equation}
\mathcal{L}_i^{\operatorname{Info}}=-\log \frac{\exp \left(z_{i} \cdot z_{i}^{\prime} / \tau\right)}{\exp \left(z_{i} \cdot z_{i}^{\prime} / \tau\right)+ \sum_{n=0}^{N} \exp \left(z_{i} \cdot z_{n}^{\prime} / \tau\right)},
\end{equation}
where $z_{i}^{\prime}$ is a positive embedding for sample $z_i$, and $z_{n}^{\prime}$ is one of the $N$ negative embeddings from other samples. $\tau$ denotes a temperature hyperparameter.
These methods all require a large amount of negative pairs. To relieve the burden, algorithms are designed with competitive performance even without negative views~\cite{grill2020bootstrap, chen2021exploring,zbontar2021barlow}. Apart from them, another line of methods, namely clustering-based methods, learn visual representations through clustering semantically similar instances in the embedding space of the models~\cite{caron2018deep, caron2020unsupervised,li2021prototypical}. 

\subsection{Lightweight Contrastive Learning}
Recent research introduces a pretrained self-supervised large model to transfer representation knowledge to small models, and greatly boost the performance of efficient models~\cite{fang2020seed,gao2021disco,xu2021bag}. In our work, we completely get rid of any help from a larger model, whilst exhibiting a competitive performance.
\subsection{Contrastive Learning with Noisy Labels}
Contrastive learning treats augmented views as positive pairs and any views from other instances as negative views. This can be unreliable, as augmented views can be semantically unrelated or distorted, and views from other instances may be semantically similar. Recent research has sought to alleviate this noise. Kalantidis et al.~\cite{kalantidis2020hard} draw inspiration from data mixing technique and mix hard negative samples in the embedding space. Chuang et al.~\cite{chuang2022robust} modify the InfoNCE loss to a symmetric one, demonstrating that it is more robust to noisy labels. Chen et al.~\cite{chen2021incremental} and Huynh et al.~\cite{huynh2022boosting} seek to identify false negatives, and then either recognize them as positives or directly remove them. In our work, we propose a simple and effective strategy to soften the one-hot distribution in the InfoNCE objective according to the hardness of negative samples. Our method is supplementary to that of Kalantidis et al. and Chuang et al., while yielding better performance than methods proposed by Chen et al. and Huynh et al.

\begin{figure*}
    \centering
    \includegraphics[width=0.98\textwidth]{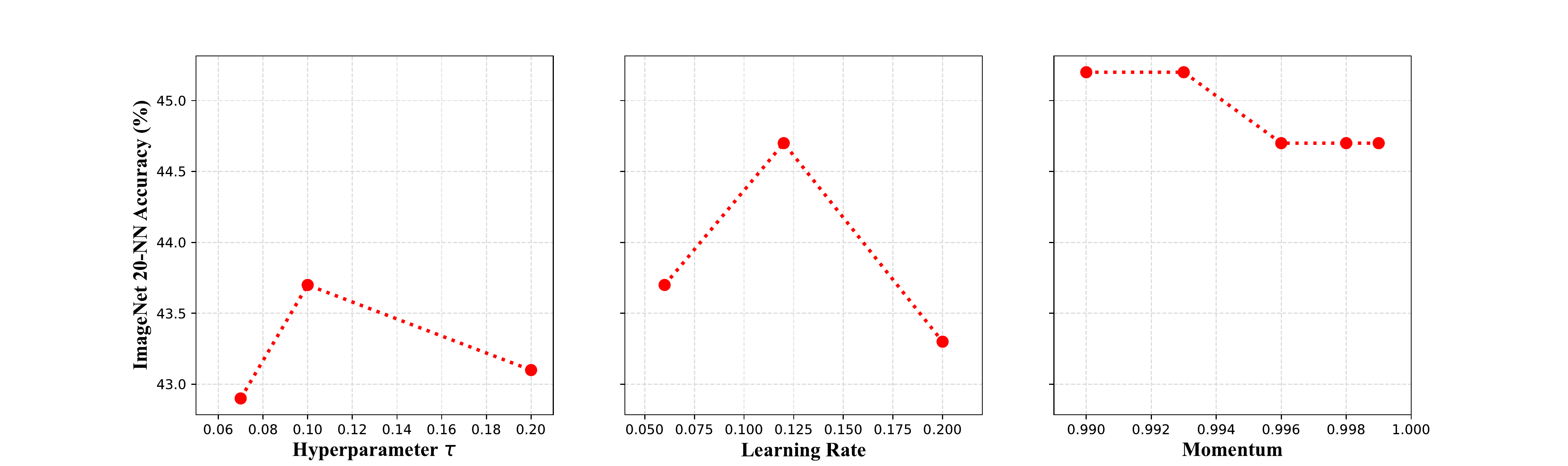}
    \caption{Hyperparameter search for ResNet18. We report 20-NN evaluation accuracy on ImageNet. The optimal result is achieved with the following setting: $\tau=0.1$, $LR=0.12$, $momentum=0.99$.}
    \label{fig:my_label}
\end{figure*}

\section{Establishing a Stronger Baseline}
In this section, we first incorporate the state-of-the-art training techniques for contrastive learning while searching for the optimal hyperparameters for lightweight models. We show the drastic improvements from these efforts. We also propose a simple reweighted loss to further boost the performance. 

We select MoCo-V2 as our backbone following the common choice~\cite{fang2020seed,gao2021disco,xu2021bag}, and this method is memory-friendly to be run on an 8-gpu server. 
\subsection{A Better Recipe for Training Lightweight SSL method}

We comprehensively search for the best training setting on lightweight models. We present an additive study in Fig.~\ref{fig:my_label} and Table~\ref{tab:hyper2}. We first tune the hyperparameters based on ResNet-18, and then verify them on EfficientNet-B0. We mainly consider the following settings:
    \subsubsection{MLP} We follow Shi et al.~\cite{shi2022efficacy}'s suggestion to adopt a wider MLP. Specifically, We set the dimension of the two-layer MLP projector to [2048,128]. On ResNet-18, we see minor improvement, but on EfficientNet-B0, the accuracy increase is drastic. This is due to the fact that previous work only add one-layer projection head on EfficientNet-B0~\cite{fang2020seed}.
    
\begin{table}
% \small
    \centering
        \caption{Results of additive study for ResNet18 and EfficientNet-B0 on ImageNet after 200-epoch self-supervised training. ``Mom.'': momentum,  ``Sym.'': symmetric. The final result shows an absolute accuracy increase of 17.9\% in total for EfficientNet-B0.}
    \label{tab:hyper2}
    \begin{tabular}{lcccc}
    \toprule
    \multirow{2}{*}[-2pt]{Setting}& \multicolumn{2}{c}{ResNet18}& \multicolumn{2}{c}{EfficientNet-B0}
     \\\cmidrule(lr){2-3}\cmidrule(lr){4-5}
      &20-NN  & $\Delta$  &Top-1  & $\Delta$ \\
        
     \midrule
      Base       & 42.9& -   &   42.2 &- \\
      \midrule
      + MLP        &43.1 & +0.2 & 53.3 & \textbf{+11.1}\\
      + Lower $\tau$&43.7&+0.6& 54.8 & +1.5 \\
      + Larger LR  &44.7&\textbf{+1.0}  & 58.4 & \textbf{+3.6}\\ 
      + Mom. &45.3& +0.6 & 57.8 & -0.6 \\
      + Symmetric Loss  &47.2& \textbf{+1.9} & 61.6 &\textbf{+3.8} \\
      + SoftNCE         &\textbf{48.8}& \textbf{+1.6}  & \textbf{65.8} &\textbf{+4.2} \\
      \bottomrule
    \end{tabular}
\end{table}
    \subsubsection{Temperature $\tau$} The temperature plays an important role in contrastive learning by controlling the gradients of hard negatives \cite{zhang2021temperature}. We test different values of $\tau$ as 0.2, 0.1 and 0.07. The selection of value 0.1 brings 0.6\% and 1.5\% absolute accuracy increase to ResNet-18 and EfficientNet-B0 separately. 
     \subsubsection{Larger learning rate (LR)} According to \cite{shi2022efficacy}, small models tend to have slow convergence in contrastive learning. An immediate thought is to enlarge the learning rate. However, we observe the training instability phenomenon when directly increasing the LR. The problem is addressed after introducing warmup technique. We linearly increase the LR from 0 to the setting value for 5 epochs. Apart from faster convergence, a larger learning rate also takes a bias towards converging to a flatter minimum, which is the key to generalize~\cite{lewkowycz2020large,li2019towards}. In our experiments, one of the most significant improvements comes along this strategy.
     \subsubsection{Momentum for the moving average encoder} MoCo updates the momentum encoder by: $\theta_{\mathrm{k}} \leftarrow m \theta_{\mathrm{k}}+(1-m) \theta_{\mathrm{q}}$, where $\theta_{\mathrm{k}}$ are the parameters of the momentum encoder $k$, and $\theta_{\mathrm{q}}$ are updated by back-propagation. A smaller $m$ enables quicker evolution of $k$, and may lead to faster convergence. We also follow the good practice to incorporate cosine momentum strategy~\cite{grill2020bootstrap}. While this strategy benefits ResNet-18, on EfficientNet-B0, it slightly decreases the accuracy.
     
     \subsubsection{Symmetric loss} In self-supervised learning, symmetric loss has demonstrated improved performance by swapping two augmented views as key and query respectively. Inspired by BYOL \cite{grill2020bootstrap}, we exchange the roles of the augmented views, and then pass them to encoder $q$ and $k$ separately to form a symmetric loss. This technique reuses the input images and further lifts the results.
     
     To summarize, we systematically test a few factors that may influence the final accuracies, and provide a better practice to train lightweight models based on MoCo-V2.

\begin{table}
     \caption{Top-1 results of lightweight models on ImageNet.}
      \label{tab:more}
  \centering
    \begin{tabular}{lcccc}
    \toprule
    Network &Eff-B0 &Eff-B1& Mobile-V3 &Res18 \\
               \midrule
    MoCo-V2    & 42.2 & 50.7   & 36.3   & 52.5 \\ 
        Reference \cite{shi2022efficacy} & 55.9 & 56.7 & 48.7 & 55.7 \\
    Light-MoCo  &\textbf{65.8}  & \textbf{66.9} &\textbf{62.3}  & \textbf{57.6} \\ 
    \bottomrule
    \end{tabular}

\end{table}

\subsection{Alleviating the negative effect of noisy labels}

\begin{figure}
         \centering
         \includegraphics[width=0.49\textwidth]{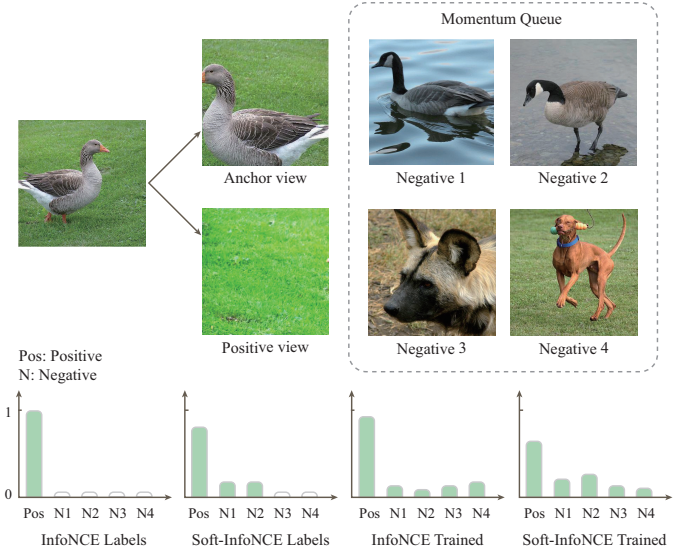}
         \caption{Illustration of noisy views and similarity distribution compared between InfoNCE and our SoftNCE. In this example, Two augmented views are generated from a given instance labelled as ``goose". The positive view is a false positive, and Negative 1 and Negative 2 are false negatives. Our SoftNCE smooths the hard labels of InfoNCE, by assigning some weight to K hardest negatives. }
          \label{image:ill}

\end{figure}

We visualize the augmented views in contrastive learning and find the common existence of noisy views, i.e., false positive views and false negative views. See the example in Figure~\ref{image:ill}. The original picture is describing “a goose on the grass”. We are interested in the \textit{goose} in the downstream task, but contrastive methods treat the augmented view about background \textit{grass} as the positive, and contrasting views including the \textit{goose} from other instances out of our expectation. To mitigate the problem, we propose a simple strategy to smooth the one-hot labels in accordance with the model's representation. 

\subsubsection{SoftNCE}

For dealing with noisy labels, smoothed labels have already been demonstrated to be more robust compared with hard labels, and label smoothing is shown to be simple yet very competitive to other techniques~\cite{song2022learning}. Inspired by this fact, we attempt to transfer the success from supervised learning to self-supervised learning with simple modification. The definition of our proposed SoftNCE is given as follows:
\begin{equation}
\mathcal{L}_i^{\operatorname{Soft}}=-{\alpha}\log \frac{\exp \left(z_{i} \cdot z_{i}^{\prime} / \tau\right)}{Z}-\sum_n{\beta_n}\log \frac{\exp \left(z_{i} \cdot z_{n}^{\prime} / \tau\right)}{Z},
\label{soft-nce-eqn}
\end{equation}
where $\alpha+\sum_n{\beta_n}=1$, and $Z$ acts as a normalizing item, $Z=\exp \left(z_{i} \cdot z_{i}^{\prime} / \tau\right)+ \sum_n \exp \left(z_{i} \cdot z_{n}^{\prime} / \tau\right)$. We will discuss different strategies of assigning values of $\alpha$ and $\beta$ as follows.

% \subsubsection{Label Smoothing without Accommodation}
% The original label smoothing technique is designed for supervised learning. Assume a cross-entropy loss between the true labels $y_t$ and the network's outputs $p_t$, as $L(y,p)=\sum_t-y_tlog(p_t)$. Label smoothing technique assign $y_t=\alpha$ for the correct class and $y_t={(1-\alpha)}/{(T-1)}$ for total T classes. In self-supervised learning, the discrimination task is performed in instance level, constituting a very large number of ``classes''. In MoCo, the size of negatives N can be very large, e.g., 65536. directly dividing the number of class leads to a value diminishing to zero.

\subsubsection{Label Smoothing with Hard Negative Mining}
Motivated by the fact that the model to some extent is able to distinguish semantic relation between instances, we propose to only smooth the top-K nearest negative instances in the momentum queue. In this way, $\beta_n=0$ if $z_{n}^{\prime} \notin \mathrm{KNN}(z_i)$. As for $z_{n}^{\prime} \in \mathrm{KNN}(z_i)$, we come up with two different patterns, the simple \textit{average pattern}, and the \textit{linear weight decay pattern}.
\begin{itemize}
    \item Average Pattern. The average pattern takes the simplest form similar to label smoothing:
    $$
     \beta_k = \frac{1-\alpha}{K},   ~z_{n}^{\prime} \in \mathrm{KNN}(z_i).
    $$w
    \item Linear Weight Decay Pattern is defined as:
$$
 \beta_k = \frac{2(K-k)}{(K-1)\cdot K}(1-\alpha),   ~z_{n}^{\prime} \in \mathrm{KNN}(z_i),
$$
where $k=1,2,...,K$, and $K>1$. This formulation satisfies $\sum_{\beta_k}=1-\alpha$, and the weight of $\beta_k$ decreases linearly with the $k^{th}$ view less similar to $z_i$. This pattern yields better performance in our experiments, and is adopted as the default setting.
\end{itemize}

\begin{table*}
\small
     \caption{Object detection and instance segmentation transferring results of ResNet-18.}
      \label{tab:detection}
  \centering
    \begin{tabular}{lcccccccccc}
    \toprule
    \multirow{2}{*}[-2pt]{Method}& \multicolumn{3}{c}{VOC Obj. Det.} & \multicolumn{3}{c}{COCO Obj. Det.} & \multicolumn{3}{c}{COCO Inst. Segm.}
    \\\cmidrule(lr){2-4}\cmidrule(lr){5-7}  \cmidrule(lr){8-10}
               &$AP^{bb}$  & $AP^{bb}_{50}$ & $AP^{bb}_{75}$ & $AP^{bb}$  & $AP^{bb}_{50}$ & $AP^{bb}_{75}$ & $AP^{bb}$ & $AP^{bb}_{50}$ & $AP^{bb}_{75}$   \\\midrule
MoCo V2  & 46.1 & 74.5 & 48.6 & 35.0 & 53.9 & 37.7 & 31.0 & 51.1 & 33.1 \\
    SEED & 46.1 & 74.8 & 49.1 &35.3 &54.2 &37.8 &31.1 &51.1 &33.2 &\\
    \midrule
    MoCo V2 (repr.)  & 50.5&77.3 & 54.4 & 35.1 &54.0 &37.7 & 31.1& 51.0& 33.3 &\\
    Light-MoCo & \textbf{51.6} & \textbf{77.9} & \textbf{56.2} & \textbf{35.4} & 54.0 & 37.8 & \textbf{31.5} & \textbf{51.3} & \textbf{33.6} \\ 
    \bottomrule
    \end{tabular}

\end{table*}

\begin{table}
     \caption{Ablation study of value $\alpha$ and $K$ in SoftNCE. We report top-1 linear probe accuracy of ResNet18 on ImageNet.}
      \label{tab:abl}
  \centering
    \begin{tabular}{lcccc}
    \toprule
    $\alpha$$\backslash$K & 5 &10& 20 & 30 \\
               \midrule
    0.6    & -  &  -  & 55.0  & - \\ 
    0.8  & 56.5  &56.6 &\textbf{57.6} & 57.2  \\ 
    0.95 & - & -& 55.6 & -\\
    \bottomrule
    \end{tabular}

\end{table}

\subsubsection{Static Smoothing vs. Incremental Smoothing}
For now, we have discussed on the value $\beta$. As for the $\alpha$, things become simpler. We propose two mechanisms. The first is just using a static $\alpha$, the second, which is more intuitive, is to increasingly smooth the labels, i.e., gradually decrease $\alpha$ with longer training, scheduled by a cosine function.  The idea is that in the initial state, the model possesses no discriminative ability, but with longer training, the model learns something useful, so more smoothing may help to resist noisy labels. Intriguingly, the experiments suggest that static smoothing performs slightly better. For incremental smoothing, the initial value of $\alpha$ is 1, and We actually tried several lower bounds of $\alpha$, 0, 0.6, 0.8, 0.95, respectively. But they all slightly perform worse than corresponding fixed $\alpha$. We assume it is because the model already learns semantic information very early after the first few epochs. We adopt the static smoothing as the default setting. We suggest setting $\alpha$ as a value between 0.7 and 0.9.

\subsection{Training Cost}
Our SoftNCE loss introduces only a little extra training cost. We find a larger momentum queue works better, and set the queue size as 256K. It costs about 3\% of the total memory. Another extra cost comes from the top-K sorting of the queue with O(N) complexity, costing less than 1\% of the total computation. Indeed, we investigate the training time, but observe no significant difference in our experiments, all around 50 hours for 200-epoch training on ResNet-18 with an 8-card V100 server.

\subsection{Hard Negative vs. False Negative}
The InfoNCE loss is shown to have a hardness-aware property, that negative examples located closer to the anchor point in the embedding space contribute more gradients during training~\cite{wang2021understanding}. That is to say, harder negatives are more useful in contrastive learning. However, if the negative samples are extremely hard, they otherwise hinder the final performance of the model. It is because that the model is able to gradually learn to capture semantic similarity during training, so the hardest negatives are likely to be semantically positive ones, i.e., false negatives. So there is a false\&hard dilemma. Previous methods dealing with the hardest negatives by directly removing them or directly treat them as positives \cite{chen2021incremental,huynh2022boosting}. We argue the importance of the smoothing technique by finding a sweet point to benefit the model to learn from hard negatives, as well as cancel the side effect from false views.

\section{Experiments}

\subsection{Settings}

% Opening
We utilize ResNet-18, EfficientNet-B0/B1, and MobileNet-V3-Large as our backbones.
Based on MoCo-V2, we deploy the SGD optimizer with learning rate as 0.2 and weight decay as 1e-4 for ResNet-18 and 1e-5 for the rest of backbones. All the networks are pre-trained on 8 Nvidia-32G-V100 GPUs with a batch size of 512 for 200 epochs, with 5-epoch warm-up. The $K$ in \textit{SoftNCE} is 20. For the momentum queue, we set the queue size as 256K, $m = 0.99$ and temperature as 0.1. We utilize a two-layer MLP projector with dimension [2048,128]. We follow the data augmentation strategy of MoCo v2.

\subsection{Linear and KNN Evaluation on ImageNet.}

Following SimSiam \cite{chen2020simple}, we finetune the pretrained networks for 90 epochs with a batch size of 4096 on ImageNet-1K. We utilize LARS as an optimizer with initial learning rate 0.1.

Results are reported in Table \ref{tab:more}. In terms of Top-1 accuracy, Light-MoCo outperforms MoCo V2 by an average of 17.72\% in accuracy. The application of SoftNCE finally reduce the performance gap between lightweight models and large models like ResNet-50.

\subsection{Transferring to Detection and Segmentation. }

Following the previous work \cite{fang2020seed}, we explore the effectiveness of Light-MoCo on two downstream tasks, object detection on VOC and instance segmentation on COCO. As shown in Table \ref{tab:detection}, Light-MoCo performs competitively on COCO and outperforms both the original reported MoCo V2 and the results reproduced by us in the settings mentioned above.

%  ImageNet 100
\subsection{Ablation Study}
We make ablation studies on the performance of our proposed SoftNCE shown in Table~\ref{tab:abl}. When $\alpha$ is close to 1, our Light-MoCo degenerates to MoCo-v2, without using any smoothing term. When $\alpha=0.95$, the performance decreases from 57.6\% to 55.6\%. We also examine our SoftNCE on a bigger model, ResNet-50 and the accuracy increase is also significant, from 67.4\% to 69.8\%.

% \section{Limitation}

% Although Light-MoCo drastically boosts the performance of lightweight contrastive models, limitations remain that we mainly test our approach on MoCo-v2, and it is still worthwhile to explore how it can be applied to other methods. Besides, Figure \ref{fig:comparison} shows that the performance of ResNet-18 is still far from ResNet-50. We leave these two limitations to future work.

% 在eff上表现很好，但resnet18表现有距离
\section{Conclusion}

Despite its thriving development, the performance of self-supervised learning suffers a dramatic fall as the size of the model decreases. In this paper, we empirically show how inadequate the existing recipe is and provide a better one for lightweight contrastive models. Besides, we propose SoftNCE, a self-supervised contrastive learning loss mitigating the side effect from noisy views. Extensive experiments show the effectiveness of our methods.

\bibliographystyle{IEEEtran}
\bibliography{icme2023template}

\end{document}